\documentclass[letterpaper, 10 pt, journal, twoside]{IEEEtran}
\usepackage[Symbolsmallscale]{upgreek}
\usepackage{xargs}
\usepackage{float}
\usepackage{wrapfig}
\usepackage[normalem]{ulem}
\usepackage{hyperref}
\usepackage{amsmath,amsfonts}
\usepackage{algorithm}
\usepackage{array}
\usepackage[caption=false,font=normalsize,labelfont=sf,textfont=sf]{subfig}
\usepackage{textcomp}
\usepackage{stfloats}
\usepackage{url}
\usepackage{verbatim}
\usepackage{graphicx}
\usepackage{cite}
\usepackage{hyperref}
\usepackage{xcolor}

\usepackage{graphics} 
\usepackage{epsfig} 
\usepackage{mathptmx} 
\usepackage{times} 
\usepackage{amsmath} 
\usepackage{amssymb}  
\usepackage{textcomp, gensymb}
\usepackage{url}
\usepackage{graphicx}
\usepackage{color}
\usepackage{xcolor}
\usepackage{soul}
\usepackage{mathtools}
\usepackage{algorithm, setspace}
\usepackage[noend]{algpseudocode}
\usepackage{ccicons}
\usepackage{comment}

\newcommand{\state}{\mathbf{x}}
\newcommand{\dyn}{f}

\newcommand{\ctrl}{u}

\newcommand{\obs}{\mathcal{O}}

\newcommand{\env}{E}
\newcommand{\policy}{\uppi}
\newcommand{\sensor}{S}
\newcommand{\image}{I}

\newcommand{\traj}[2]{\zeta^{#1}_{#2}}

\newcommand{\brt}{\mathcal{V}}
\newcommand{\ibrt}{\mathcal{I}_{unsafe}}
\newcommand{\inprod}[2]{{\left\langle #1, #2 \right\rangle}}

\begin{document}

\title{Discovering Closed-Loop Failures of Vision-Based Controllers via Reachability Analysis}

\author{Kaustav Chakraborty$^{1}$, Somil Bansal$^{1}$
\thanks{$^{1}$Authors are with the Department of Electrical and Computer Engineering, University of Southern California, Los Angeles, CA 90089, USA {\tt\footnotesize \{kaustavc, somilban\}@usc.edu.} This research is supported in part by the DARPA Assured Neuro Symbolic Learning and Reasoning (ANSR) program and by the NSF CAREER program (2240163).}
\thanks{Project Website: \url{http://vatsuak.github.io/failure-detection}}}


\maketitle

\begin{abstract}
Machine learning driven image-based controllers allow robotic systems to take intelligent actions based on the visual feedback from their environment.
Understanding when these controllers might lead to system safety violations is important for their integration in safety-critical applications and engineering corrective safety measures for the system.
Existing methods leverage simulation-based testing (or falsification) to find the failures of vision-based controllers, i.e., the visual inputs that lead to closed-loop safety violations.
However, these techniques do not scale well to the scenarios involving high-dimensional and complex visual inputs, such as RGB images. 
In this work, we cast the problem of finding closed-loop vision failures as a Hamilton-Jacobi (HJ) reachability problem. 
Our approach blends simulation-based analysis with HJ reachability methods to compute an approximation of the backward reachable tube (BRT) of the system, i.e., the set of unsafe states for the system under vision-based controllers. 
Utilizing the BRT, we can tractably and systematically find the system states and corresponding visual inputs that lead to closed-loop failures.
These visual inputs can be subsequently analyzed to find the input characteristics that might have caused the failure.
Besides its scalability to high-dimensional visual inputs, an explicit computation of BRT allows the proposed approach to capture non-trivial system failures that are difficult to expose via random simulations.
We demonstrate our framework on two case studies involving an RGB image-based neural network controller for (a) autonomous indoor navigation, and (b) autonomous aircraft taxiing.

\end{abstract}


\section{Introduction}
\label{sec:intro}
\IEEEPARstart{R} ECENT advances in computer vision and deep learning have enabled autonomous systems to employ vision-based controllers to perceive their environment and react to it for accomplishing various tasks, including agile navigation \cite{loquercio2019deep}, manipulation \cite{wang2019learning}, autonomous driving \cite{grigorescu2020survey}, and autonomous aircraft landing and taxiing \cite{katz2021verification}. 
Despite their success, such vision-based controllers can fall prey to issues when they are subjected to inputs that are scarcely encountered in the training dataset or are outside the training distribution.  
For example, a visual policy trained exclusively with well-illuminated images might fail to predict good actions in dark conditions; an autonomous car that is predominately shown to take right turns in an expert dataset, may fail to learn to make left turns. Such vision failures can cascade to catastrophic system failures and compromise system safety.
Thus, to successfully adopt vision-based neural network controllers in safety-critical applications, it is vital to analyze them and understand when and why they result in a system failure. 
In addition to reasoning about system safety, these failure modes might be useful in engineering corrective measures for the system.

While techniques from adversarial learning and robust optimization have been used to find ``adversarial'' inputs for vision components, they tend to focus on the component-level safety analysis, i.e., detecting failures or errors only within the vision component, ignoring their effect on the downstream system and the overall robot safety. 
Indeed, not all vision failures are equal from the robot safety standpoint. 
For instance, the same prediction error by a visual policy for a high-speed drone can be much more catastrophic near a wall compared to an empty hallway. Thus, it is imperative that we analyze these vision modules in conjunction with the robot dynamics. To that end, formal verification techniques have been used for system-level (or closed-loop) safety analysis of dynamical systems; however, their direct application to vision-based controllers remains impractical due to these controllers' high-dimensional and complicated input spaces and the lack of mathematical models relating the robot state to the visual input at that state. 
Simulation-based testing has been a promising approach to overcoming these challenges; by treating the system as a black box, one can search for system trajectories that result in a failure under the vision-based controller. 
However, this process (also called falsification) can be highly time-consuming, and it struggles with exposing long-tail of system failures. 

In this work, we cast the problem of finding closed-loop vision failures as a Hamilton-Jacobi (HJ) reachability problem and compute the Backward Reachable Tube (BRT) of the system. 
Given a set of unsafe states (e.g., obstacles for a navigation robot), the BRT is the set of all starting states of the system which ultimately reach an unsafe state under the vision-based controller. 
Thus, the BRT captures all possible unsafe states of the system. 
The sequences of visual inputs corresponding to the states in the BRT can be, therefore, classified as the inputs that result in closed-loop system failures, enabling us to discover failures in a systematic manner. 
Finding closed-loop failures with the help of a BRT also circumvents the need for a direct search in the high-dimensional input space, leading to a tractable discovery of the closed-loop failures.

Typically, the BRT computation requires an analytical model of closed-loop system dynamics. However, such a model is challenging to obtain for vision-based controllers due to the lack of a mathematical model between the system state and the corresponding visual input (and, by extension, the control input at that state). 
Our key idea in overcoming this challenge is to blend level set-based reachability methods with simulation-based methods to compute a numerical approximation of the BRT. 
Level set methods compute the BRT over a state-space grid; even though level set methods have typically been employed in the settings where an analytical model of the closed-loop dynamics is available, they only need the system dynamics at the state grid points. 
Motivated by this observation, we leverage readily available photo-realistic simulators to obtain the visual inputs corresponding to the state grid points and, subsequently, the control inputs. This allows us to compute (approximate) BRTs under a vision-based controller without knowing an analytical model of the environment but rather only from its samples. Our approach is particularly suitable for the vision-based controllers that are trained in simulation or where simulators are readily available for testing.

In summary, the major contributions of this paper are,
\begin{itemize}
    \item We propose a framework that bridges formal guarantees of HJ reachability with simulation-based methods for a closed-loop safety analysis under vision-based controllers.
    \item We demonstrate the generalizability of our method by studying two distinct systems, which involve a 3D and a 5D dynamical system using RGB image-based neural network controllers.
    \item We analyze the obtained image sequences to deduce common failure ``scenarios'' for the system.
\end{itemize}   

\section{Related Work}
\label{sec:related}
\noindent \textbf{Component-level safety analysis.} A number of recent works have proposed tools to formally verify the input-output properties of neural networks (NN) \cite{tjeng2017evaluating,katz2017reluplex,pmlr-v151-brown22b}. Even for image-inputs, there have been significant advances in generating adversarial input perturbations \cite{huang2017safety} that result in erroneous outputs, as well as analysis of neuron activations \cite{pei2017deepxplore} to different input images. 
However, finding the failures of the vision-module in isolation from the rest of the system does not address the closed-loop system safety challenges.

\vspace{0.2em}
\noindent \textbf{Closed-loop verification.} Closed-loop verification techniques combine the output of input-output NN verification methods with reachability analysis to provide guarantees on the closed-loop performance of NN controllers \cite{huang2019reachnn, xiang2018output, julian2019guaranteeing}.
While these approaches work well for state-based controllers with low-dimensional input
spaces, they do not scale to high-dimensional vision-based controllers.
An additional challenge with perception inputs is that it is challenging to even define the
observation space for verification – for example, not all 256 × 256 × 3 arrays make a valid real-world RGB image. 
To overcome these challenges, there have been attempts to abstract the observation
space through GANs \cite{katz2021verification}, piecewise affine abstractions \cite{hsieh2021verifying}, or
a geometric sensor mapping \cite{santa2022nnlander}.
However, the obtained failures are only as accurate as the abstraction itself.
Moreover, obtaining accurate abstractions for complex real-world image inputs can be challenging.

\vspace{0.2em}
\noindent \textbf{Closed-loop falsification.} On the other hand, the development and wide availability of photo-realistic simulators and datasets \cite{xplane, matterport, SBPD} has presented an opportunity to use them as a black-box representation between the system state and the corresponding visual input, without requiring an explicit analytical model of robot sensor. 
This has led to development of approaches that find closed-loop perception failures through forward simulation (also referred to as falsification) \cite{indaheng2021scenario,fremont2020formal}.
Even though promising, these approaches often rely on heuristics to effectively search over the image space for finding the failures, including low-dimensional feature encoding and adaptive Bayesian sampling in the feature space \cite{dreossi2017systematic,ghosh2021counterexample}. 
However, it is not immediately apparent how to obtain informative low-dimensional feature encoding for complex RGB images.
In addition, forward simulation approaches can be computationally prohibitive for finding rare, long-tail failure cases.
To overcome this challenge, \cite{yang2021synthesis} leverages linear system dynamics along with a simulator to efficiently search for closed-loop failures. 
Our work builds upon this line of work to integrate sampling-based falsification approaches with HJ reachability analysis for exposing closed-loop failures of vision-based controllers for general non-linear systems.

\section{Problem Setup}
\label{sec:problem}
Let us consider a robot in an environment, $\env$. 
The environment broadly refers to all the factors that are external to the robot (e.g. buildings in which a robot is navigating, the goal/obstacle location, or even characteristics like different weather conditions, time of the day, or camera parameters that might effect the vision of the robot).  

We model the robot as a dynamical system with state $\state\, \in \mathbb{R}^n$, control $\ctrl\, \in \mathcal{U}$(a compact set), and dynamics:
\vspace{-0.4em}
\begin{equation}
\label{eqn:dyn_gen}
    \dot{\state} = \dyn(\state, \ctrl)
    \vspace{-0.4em}
\end{equation}
Let $\sensor$ denote the robot’s sensor mapping from a state $\state$ to an output (or observation), e.g., a depth or an RGB image $I = \sensor(\state, \env)$.
In this work, we specifically focus on vision sensors for which $I$ is often high-dimensional.
Even though obtaining an analytical model of $S$ is non-trivial for vision sensors, we assume access to a simulator that allows us to query $I$ for a state $\state$.
This allows us to leverage recent advances in photorealistic simulators to find vision failures.
 
Let $\policy$ denote an output-feedback control policy, that maps observations to the control input $\ctrl$:
\vspace{-0.4em}
\begin{equation}
    \ctrl \coloneqq \policy(I, \state, \env)
    \label{eqn:control_gen}
    \vspace{-0.4em}
\end{equation}
For vision-based policies, $\policy$  often involves neural networks. 
In our work, $\policy$, could be a single end-to-end learning model or can be composed of different submodules. 
For example, in robot navigation, $\policy$ might consist of a learning-based visual route planner that serves as an initial guess for an optimization-based trajectory planner for robot control.
Let $\traj{\policy}{\state}(\tau)$  be the robot's state achieved at time $\tau$ when it starts at state $\state$ at time $t=0$, and follows the policy $\policy$ over $[0,\tau]$.
 Finally, let us denote a set of undesirable or unsafe states by $\obs \subset \mathbb{R}^n$.
 For example, $\obs$ could represent obstacles for a navigation task, or off-runway positions for an autonomous aircraft.

In this work, we are interested in finding sequences of input images that lead to a closed-loop system failure. 
In other words, we wish to find the set of images $\ibrt$, which, when seen by the visual controller, leads the system into $\obs$. We hypothesize that analyzing $\ibrt$, will then uncover particular properties of the visual inputs that cause the robotic system to fail. 

Note that we are interested in finding the visual inputs that lead to a closed-loop failure of the system and not just that of the vision module. 
However, finding such closed-loop failures is challenging due to (a) high-dimensional observations $\image$, and (b) the lack of an analytical model of $\sensor$. Our approach of blending simulation-based techniques with HJ-reachability analysis is a key contribution towards overcoming these challenges.

\vspace{0.2em}
\noindent \textbf{\textit{Running example (TaxiNet).}} Now we introduce the aircraft taxiing problem \cite{katz2021verification} that we will use as a running example to illustrate the key concepts.
Here, the robot is a Cessna 208B Grand Caravan modelled as a three-dimensional non-linear system with dynamics:
\vspace{-0.8em}
\begin{equation}
\label{eqn:dyn_taxinet}
     \dot p_x=v\, cos(\theta)\quad \dot p_y=v\, sin(\theta)\quad  \dot \theta = u
     \vspace{-0.6em}
\end{equation}
where $p_x$ is the crosstrack error (CTE), $p_y$ is the downtrack position (DTP)
and $\theta$ is the heading error (HE) of the aircraft in degrees from the centreline (Fig. \ref{fig:views_reflection}(a) shows how these quantities are measured). $v$ is the linear velocity of the aircraft kept constant at 5 m/s, and the control $u$ is the angular velocity. 

The goal of the aircraft is to follow the centreline as closely as possible using the images obtained through a camera mounted on its right wing. 
For this purpose, the aircraft uses a Convolutional Neural Network (CNN), which returns the estimated  CTE, $\hat p_x$, and the estimated HE, $\hat \theta$.
A proportional controller (P-Controller) then takes these predicted tracking errors to return the control input, as follows:
\vspace{-0.6em}
\begin{equation}
    \label{eqn:pctrl}
    \ctrl \coloneqq tan(-0.74\hat p_x -0.44\hat \theta)
    \vspace{-0.6em}
\end{equation}
Hence, the policy $\policy$ is a composition of the CNN and the P-Controller.
Intuitively, the P-controller is designed to steer the aircraft towards the centreline based on the state estimate provided by the CNN.

The image observations are obtained using the X-Plane flight simulator that can render the RGB image, $I$, from a virtual camera ($\sensor$) mounted on the right wing of the aircraft at any state and a given time of the day (see Fig. \ref{fig:views_reflection}(c), (d) for representative images).
Note that the CNN here is also trained in simulation using the data collected from X-Plane. Please see \cite{katz2021verification} for the training details.

We define the unsafe states for the aircraft as $\mathcal{O} = \{\state: |p_x| \geq 10\}$, which corresponds to the aircraft leaving the runway.
The environment $\env$ is the runway 04 of Grant County International Airport. 
Our goal is to find the set of input images that eventually drive the aircraft off the runway under the control policy in \eqref{eqn:pctrl}.

\section{Background: Hamilton-Jacobi Reachability}
\label{sec:background}
In this work, we will find closed-loop system failures using HJ reachability analysis. In reachability analysis, one is interested in computing the Backward Reachable Tube (BRT) -- the set of all initial states, such that an agent starting from those states will reach the target set $\obs$ within the time horizon $[t,T]$ under policy $\policy(\state)$. 

We define the BRT for a closed loop system as follows:
\vspace{-0.6em}
\begin{equation}
\vspace{-0.6em}
    \mathcal{V} \coloneqq \{\state: \exists \tau \in [t,T], \zeta^{\policy}_{\state}(\tau) \in \mathcal{O} \}
    \label{eqn:brt}
\end{equation}
HJ reachability allows us to compute the BRT for general nonlinear systems, while handing cases of control and disturbance inputs to the system over arbitrary shaped target sets. 
In HJ reachability, the BRT can be computed using level-set methods \cite{mitchell2005time, mitchell2002level}. 
In the level-set methods, the target set is represented as a sub-zero level set of a function $l(\state)$, i.e., $\obs = \{\state: l(\state) \leq 0\}.$ 
Usually, $l(\state)$ is given by the signed distance function to $\obs$.
With this formulation, the BRT computation can be reframed as an optimal control problem that requires the computation of a value function defined as:
\vspace{-0.6em}
\begin{equation}
    V(\state,t) = \min_{\tau \in [t,T]}l(\zeta^{\policy}_{\state}(\tau))
    \vspace{-0.6em}
    \label{eqn:vfn}
\end{equation}
The value function in \eqref{eqn:vfn} can be computed recursively using the dynamic programming principle. This results in a partial differential equation referred to as Hamilton-Jacobi-Bellman Variational Inequality (HJB-VI) \cite{mitchell2005time}:
\vspace{-0.5em}
\begin{equation}
\begin{aligned}
    \min\{D_tV(\state,t)+H(\state,t)&,l(\state) - V(\state,t)\}=0\\
    \text{with }V(\state,T) = l(\state)
    \label{eqn:hjivi}
    \vspace{-0.6em}
\end{aligned}
\end{equation}
Here, $D_t$ and $\nabla$ represent the time and spatial gradients of the value function. 
$H \coloneqq \inprod{\nabla V(\state,t)}{f(\state, \policy(\state)}$ is the called the Hamiltonian. Intuitively, \eqref{eqn:hjivi} is a continuous-time counterpart of the Bellman equation in discrete time. Once the value function is computed, the BRT is given as the set of states from which entering into the target set cannot be avoided, i.e., the optimal signed distance to $\obs$ is negative. Thus, the BRT is given by the subzero level set of the value function:
\vspace{-0.5em}
\begin{equation}
    \mathcal{V} = \{\state: V(\state,t) \leq 0\}
    \vspace{-0.5em}
\end{equation}
 
In the next section, we discuss how $\mathcal{V}$ can be used to find closed-loop failure inputs for vision-based policies.

\section{Closed-Loop Failure Discovery via HJ Reachability}
\label{sec:approach}
We cast the problem of finding closed-loop vision failures as a HJ reachability problem. 
Specifically, given the set of undesirable states $\obs$, the sensor mapping can be composed with the vision-based controller to obtain the closed-loop, state-feedback policy, $\hat\policy$ for a given environment:
\vspace{-0.65em}
\begin{equation}
    u = \policy(\image,\state,\env)\;=\;\policy(\sensor(\state,\env),\mathbf{x},E)\;
    \implies \ctrl = \hat\policy(\state)\quad
    \vspace{-0.65em}
\label{eqn:ctrl_state}
\end{equation}
Given the policy $\hat\policy$, we compute the BRT $\brt$ by solving the HJB-VI in \eqref{eqn:hjivi}.
Intuitively, $\brt$ represents the set of all initial states of the robot that will eventually enter $\obs$ under the vision-based controller.
Once computed, $\brt$ can be used to obtain the failure inputs as:
\vspace{-0.7em}
\begin{equation}
\vspace{-0.7em}
    \ibrt =\{I: I = S(\mathbf{x},E), \mathbf{x} \in \mathcal{V}\},
    \label{eqn:problem}
\end{equation} 
which can be subsequently analyzed to find the common characteristics of the failure inputs.
Finding $\ibrt$ through a BRT allows us tractably search for failures over the high-dimensional input (image) space by converting it into a search over the state space, which is typically much lower dimensional.

However, existing HJ reachability methods typically require an analytical expression of $\hat\policy$ to solve the HJB-VI, which in turn requires an analytical model of the sensor mapping $S$. 
Unfortunately, obtaining such models of $\sensor$ remains a challenging problem, especially for visual sensors; consequently, obtaining closed-loop policy $\hat\policy$ in an analytical form is difficult. 
To overcome this challenge, we leverage level set methods \cite{mitchell2007toolbox} to compute a numerical approximation of the BRT.
Level set methods solve the HJB-VI numerically over a uniformly discretized state-space grid. 
Thus, it is sufficient to know the output of $\hat\policy$ at the discretized states.
To compute the controller output at the grid points, we turn to photorealistic simulators that allow us to query $I$, and subsequently $\ctrl$, at any state $\state$.
Thus, blending simulator-based sampling with level set methods from HJ reachability allows us to approximate the BRT from the samples of $\sensor$, in place of a rigorous model of $S$.
The approximate BRT can then be used to find the failure inputs as in \eqref{eqn:problem}.

In summary, our proposed approach consists of three main steps: \textbf{(1)} obtaining the closed-loop dynamics of the system via sampling over a uniform grid of the robot's statespace, \textbf{(2)} computing the BRT to extract the set of images leading to closed-loop system failures, and \textbf{(3)} analyzing these images to uncover their interesting properties.
Overall, the proposed approach enables the failure analysis of complicated and high-dimensional visual controllers without possessing an explicit analytical model of the environment or the sensor.

\section{Case Studies}
\label{sec:cases}
\subsection{Autonomous Aircraft Taxiing (TaxiNet)}
For the running example, we obtain the BRT over the statespace: $[-11,11]m$ for $p_x$, $[100,250]m$ for the $p_y$ and $[-28\degree, 28\degree]$ for $\theta$. 
In order to compute the BRT we use a uniform $101\times101\times101$ grid along $p_x$, $p_y$ and $\theta$ directions and leverage the Level Set Toolbox \cite{mitchell2007toolbox} to solve the HJB-VI numerically. 

\begin{figure}[htb!]
\centering
\vspace{-0.75em}
\includegraphics[width=\columnwidth]{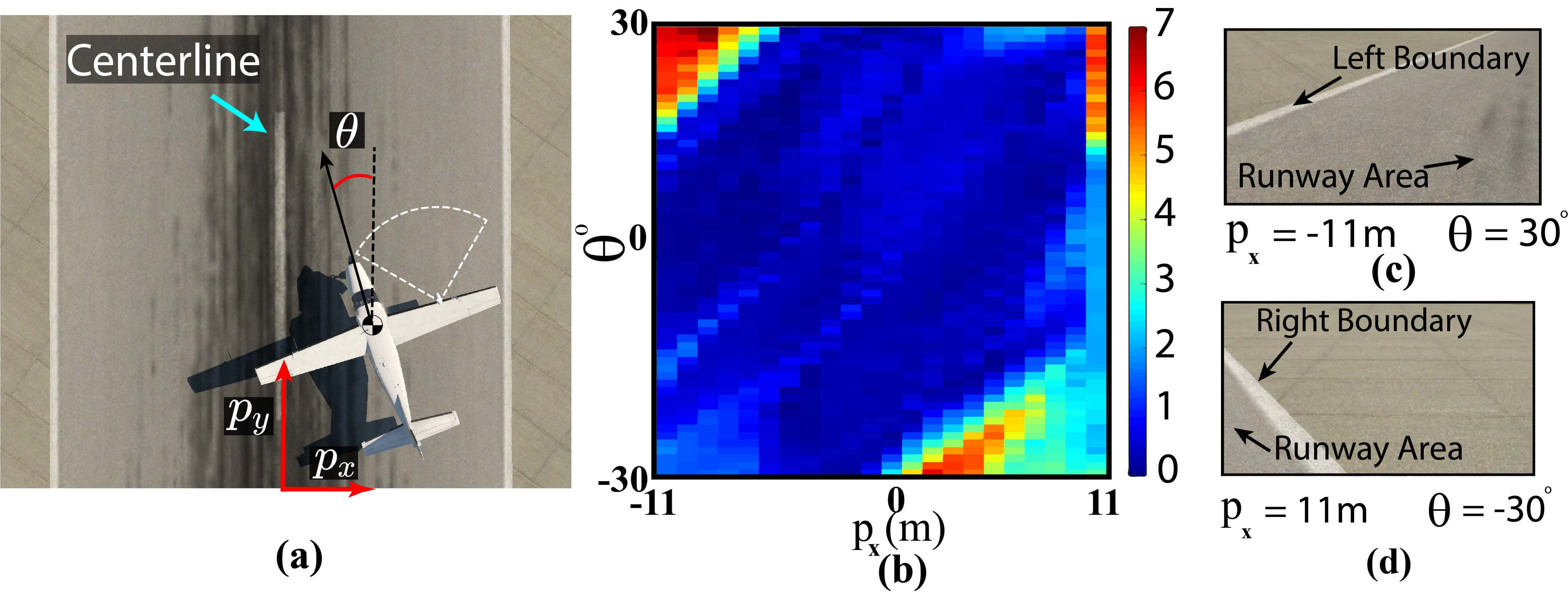}
\vspace{-2.0em}
\caption{\textbf{(a)} The autonomous aircraft taxiing example. $p_x$, $p_y$, $\theta$\ denote the state of the aircraft; the FoV of the camera is shown with dashed-white lines. \textbf{(b)} Heatmap showing the error in prediction over $p_x\text{ and }\theta$ for $p_y=160m$.  Asymmetric images \textbf{(c), (d)} seen by the CNN at symmetric locations about the centerline. 
}
\vspace{-0.75em}
\label{fig:views_reflection}
\end{figure}

To obtain $\hat{\policy}(\state_i)$ at any grid point $\state_i$, we obtain the $360 \times 200 \times 3$ RGB image observation from the virtual camera ($\sensor$) in the X-plane simulator (see Figs. \ref{fig:views_reflection}(c),(d).
This image is fed through the CNN ($C$) to obtain the estimates of crosstrack and heading errors, which are subsequently utilized by the P-controller ($P$) in \eqref{eqn:pctrl} to obtain the control input.
Hence $\hat{\policy}$ for this study is the composition, $P\circ C \circ S$.
\begin{figure}[h!]
    \centering
    \vspace{-0.7em}
         \includegraphics[width=0.85\columnwidth]{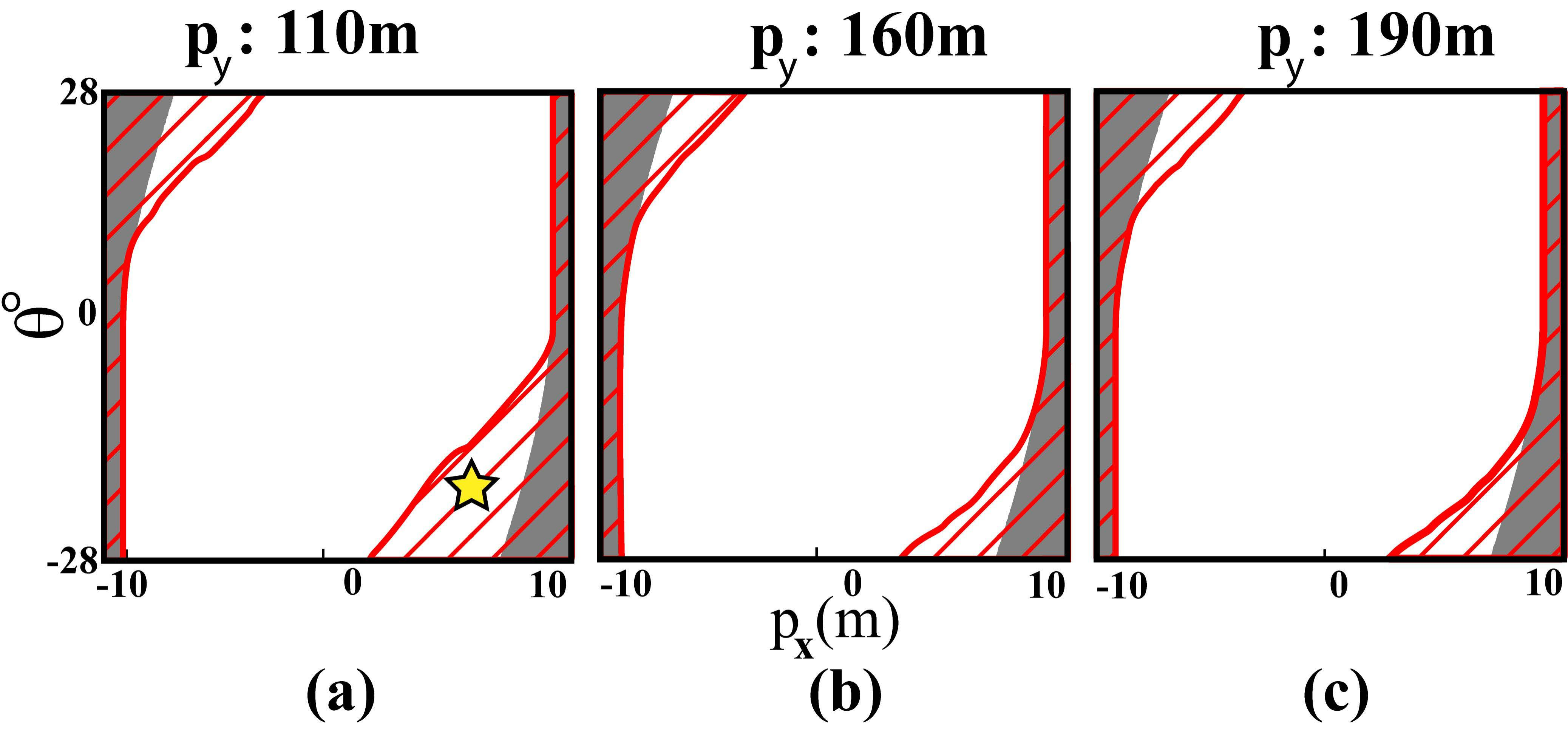}
         \vspace{-0.7em}
         \caption{Closed-loop BRTs of the ideal system (the grey area) and the actual system (the striped red area) during the morning when the aircraft starts at $p_y$ \textbf{(a)} 110m, \textbf{(b)} 160m, and \textbf{(c)} 210m. 
         The vision-based controller leads to particularly unsafe behaviors near the boundary of the runway.
         }
         \vspace{-0.75em}
         \label{fig:morning_ys}
     \end{figure}

The slices of the obtained BRT (the red striped area) are shown in Fig. \ref{fig:morning_ys} for different values of downtrack position $p_y$. With the change in $p_y$, the CNN is able to observe different regions along the length of the runway which effects its prediction capabilities. The ability of the aircraft in successfully accomplishing its taxiing task (reflected through the area enclosed by the BRT) was hence seen to be critically dependent on the starting $p_y$.
For comparison purposes, we also compute the system BRT assuming an ``ideal'' CNN (the shaded grey region). 
In this case, it would mean that the CNN predicts the system state perfectly (referred to as the ideal system henceforth). 
Any observed performance drop from this assessment on reintroducing the CNN in the pipeline (called the actual system henceforth) can be attributed to the vision module.
Note that, as expected, the BRT for the ideal system is a proper subset of the BRT for the actual system.
We now leverage the obtained BRT to analyze different failure modes of the vision-based controller.

\noindent \textbf{Failures near the boundary of the runway}.
Upon comparing the BRTs for the actual and the ideal systems, it is evident that the vision module particularly leads to more safety violations near the boundary (left and right) of the runway.
To analyze these failures further, we plot the error heatmap between the predicted state (by the CNN) and the actual state for different $p_x$ and $\theta$ for $p_y = 160m$ in Fig. \ref{fig:views_reflection}(b). We observe that the top-left and the bottom-right areas of the heatmap display an unusually high localization error by the CNN. On querying some representative images (\ref{fig:views_reflection}(c),(d)) observed by the CNN at these critical states, we find that when the aircraft is closer to the boundary of the runway while looking away from the center lane, only a reduced portion of the runway is visible in the image observations, leading to erroneous predictions from the CNN. Such errors, in turn, lead to unsafe control decisions, leading to a striking resemblance between the BRT (Fig. \ref{fig:morning_ys}) and the heatmap (Fig. \ref{fig:views_reflection}(b)).

\noindent \textbf{Failures due to asymmetric camera placement}.
It is also interesting to note that even though the BRT is symmetric for the ideal system, it is not the case for the actual system (Fig. \ref{fig:morning_ys}). 
Specifically, the size of the BRT is bigger when the vehicle is to the right of the centreline.
Upon a closer inspection, it turns out that the camera is mounted on the right wing of the aircraft, leading to asymmetric observations from the two sides of the centreline, as shown in Fig. \ref{fig:views_reflection}(c), (d). 
This leads to the aircraft having a wider view of the runway when it starts from the left of the centreline (Fig. \ref{fig:views_reflection}(c) has more view of the track than Fig. \ref{fig:views_reflection}(d)), leading to a better localization by the CNN.
This example demonstrates the efficacy of the proposed approach in comparing the failures of the controller in different parts of the environment.

\noindent \textbf{Failure due to the presence of runway marking}. We analyze the changes in the BRT with variations in the starting positions of the aircraft along the runway, i.e., with $p_y$. 
We observe that the BRT corresponding to $p_y$ = 110m (Fig. \ref{fig:morning_ys}(a)) is peculiarly large, especially around the bottom-right (when the aircraft starts on the right of the runway with a negative heading).
\begin{figure}[h!]
    \centering
    \vspace{-0.75em}
         \includegraphics[width=0.85\columnwidth]{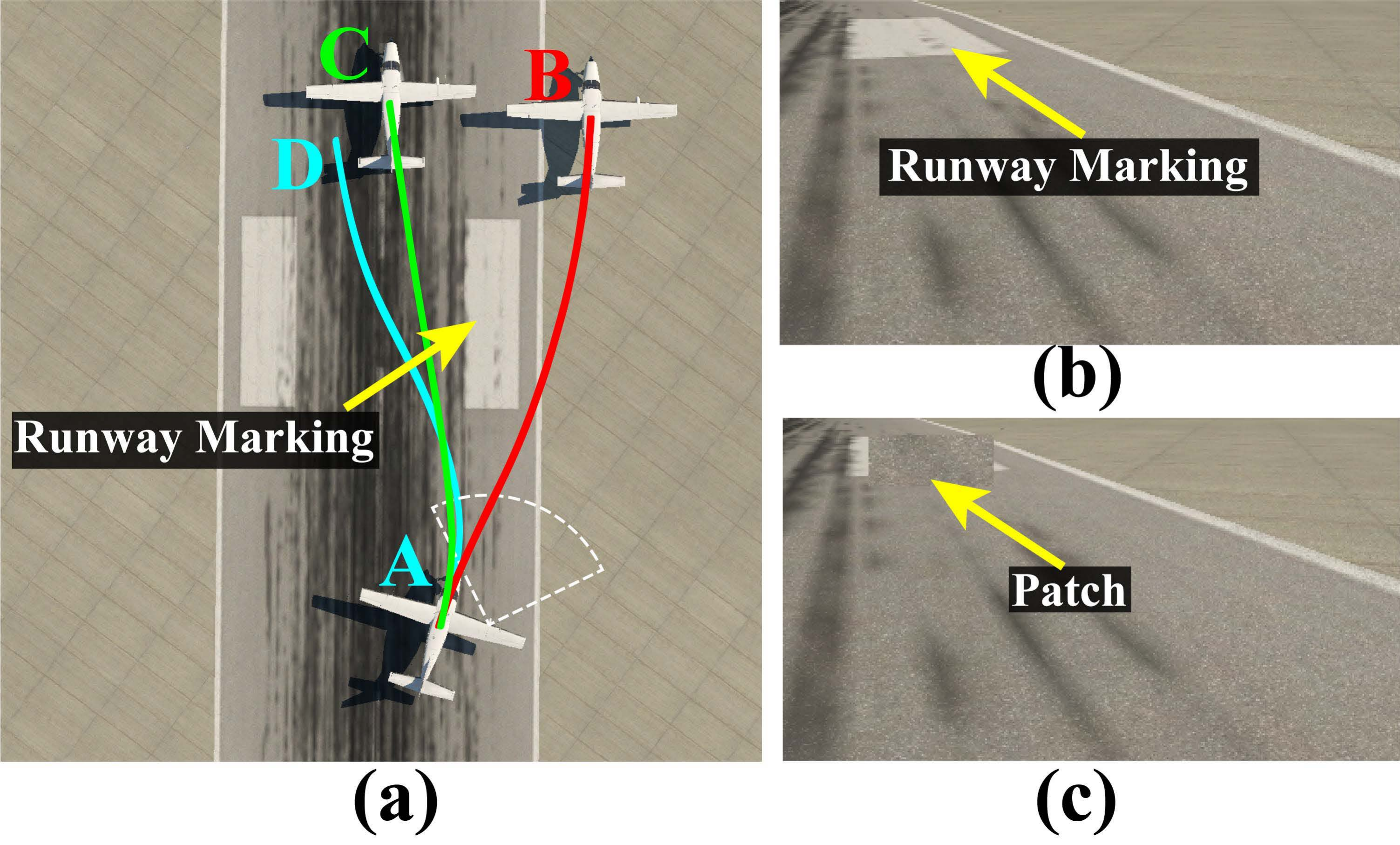}
         \caption{\textbf{(a)} Top-view of the runway in the morning. The trajectory followed by the aircraft under the CNN policy (red line) takes it off the runway. The successful trajectory (in green) takes the aircraft from ``A" to ``C", on adding the patch over the runway marking during ablation. The trajectory (in cyan) from ``A" to ``D" is followed at night. 
         \textbf{(b)} The actual image that the CNN sees at ``A" (yellow star in Fig. 3(a)). The CNN confuses the runway marking as the centreline. \textbf{(c)} Modified image with an artificial patch over the runway marking.}
         \vspace{-0.5em}
         \label{fig:pedcross_fail_morning}
\end{figure}
We simulate the aircraft trajectory from a state in the BRT (marked with the yellow star in Fig. \ref{fig:morning_ys}(a)) and query the images observed by the aircraft along the trajectory. 
The aircraft trajectory is shown in red in Fig. \ref{fig:pedcross_fail_morning}(a).
We notice that the CNN confuses a runway side strip marking (Fig. \ref{fig:pedcross_fail_morning}(b)) with the centreline and steers the aircraft towards it, ultimately leading the aircraft into the unsafe zone ($|p_x|$ $\geq$ 10m). We also perform an ablation study where we mask off the area of the runway marking in the morning with a patch having the same color as the runway (shown in Fig. \ref{fig:pedcross_fail_morning}(c)).
The corresponding aircraft trajectory is shown in green in Fig.\ref{fig:pedcross_fail_morning}(a).
On being unable to see the runway marking, the aircraft completed its taxiing task successfully, indicating that the runway marking is indeed ``fooling" the CNN.
This example illustrates that the shape and analysis of the BRT makes it easier to find such subtle failures of the vision-based controller, which are hard to expose via random sampling.
The two methods are compared further toward the end of this section.

\noindent \textbf{Effect of the time of the day on the vision failures}.
To understand the impact of the time of the day on the vision failures, we also compute the BRT of the system at the night time (2100 hrs.). 

\begin{figure}[h!]
    \centering
\includegraphics[width=0.85\columnwidth]{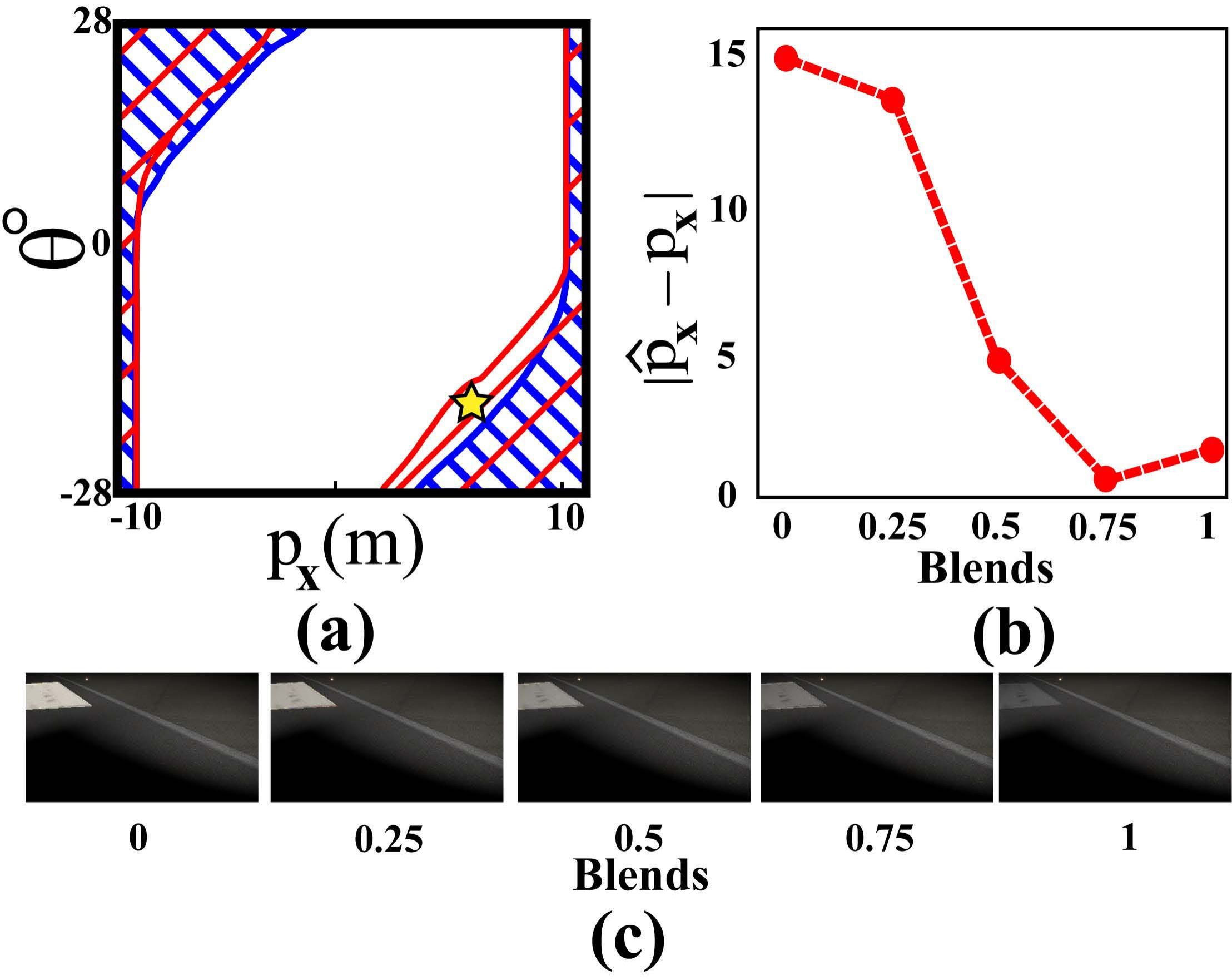}
    \caption{
    \textbf{(a)} The morning (red shaded), and the night (blue shaded) BRTs overlaid for a $p_y$ of 110m. The state of interest, shown with a yellow star, is only contained in the morning BRT and not in the night BRT. \textbf{(b)} The absolute difference in $\hat{p_x}$ and the ground truth $p_x$ vs. the different blends of the runway marking at the state of interest (a lower value is better). \textbf{(c)} The images (at the yellow star in (a)) corresponding to different blends. The right-most image has a blend of 1, which is the unmodified image that the aircraft observes at night. 
    The left-most image with a blend of 0 is obtained by manually cropping the runway marking and replacing it by the patch from the image observed in the morning. 
    Any intermediate image is an interpolation between these two according to the blend value.}
         \label{fig:pedcross_night}
         \vspace{-1.5em}
     \end{figure}

In Fig. \ref{fig:pedcross_night}(a), we overlay the BRT for the morning and the night time for $p_y = $110m. 
The overlaid BRTs provide us a quick way to systematically check the CNN's performance for all states of interest.
 We saw that the BRT for the night was smaller than that for the morning (around the bottom-right area), indicating that during the night time the CNN was not affected by the runway marking. 
 This was confirmed via the trajectory followed by the aircraft at night (the cyan trajectory in Fig. \ref{fig:pedcross_fail_morning}(a)). 
 It was a surprising discovery since night-time performance is typically worse for vision-based systems due to reduced visibility and illumination. 
 However, it seems that the aircraft benefits from not being able to see certain parts of the runway that can potentially confuse its vision system. 
 In this case, the CNN could not see the runway marking at night due to low illumination, obviating a possible confusion with the centreline! 
 In order to understand the effect of the visibility of runway marking on the failure of the CNN's prediction, we feed the CNN with different night-time images at the same state (Fig. \ref{fig:pedcross_night}(c)). 
 However, in each of these images we blend (with varying proportions) a cropped-out version of the runway marking that the CNN sees at the same state in the morning. 
 This changes the effective illumination/brightness only around the marking while keeping the rest of the image the same. 
 We found that the error in the $p_x$ prediction decreased on decreasing the runway marking brightness (Fig. \ref{fig:pedcross_night}(b)), showing that the CNN is indeed able to make correct predictions when the marking is not well illuminated.
   
\begin{figure}[ht]
\centering
     \includegraphics[width=0.85\columnwidth]{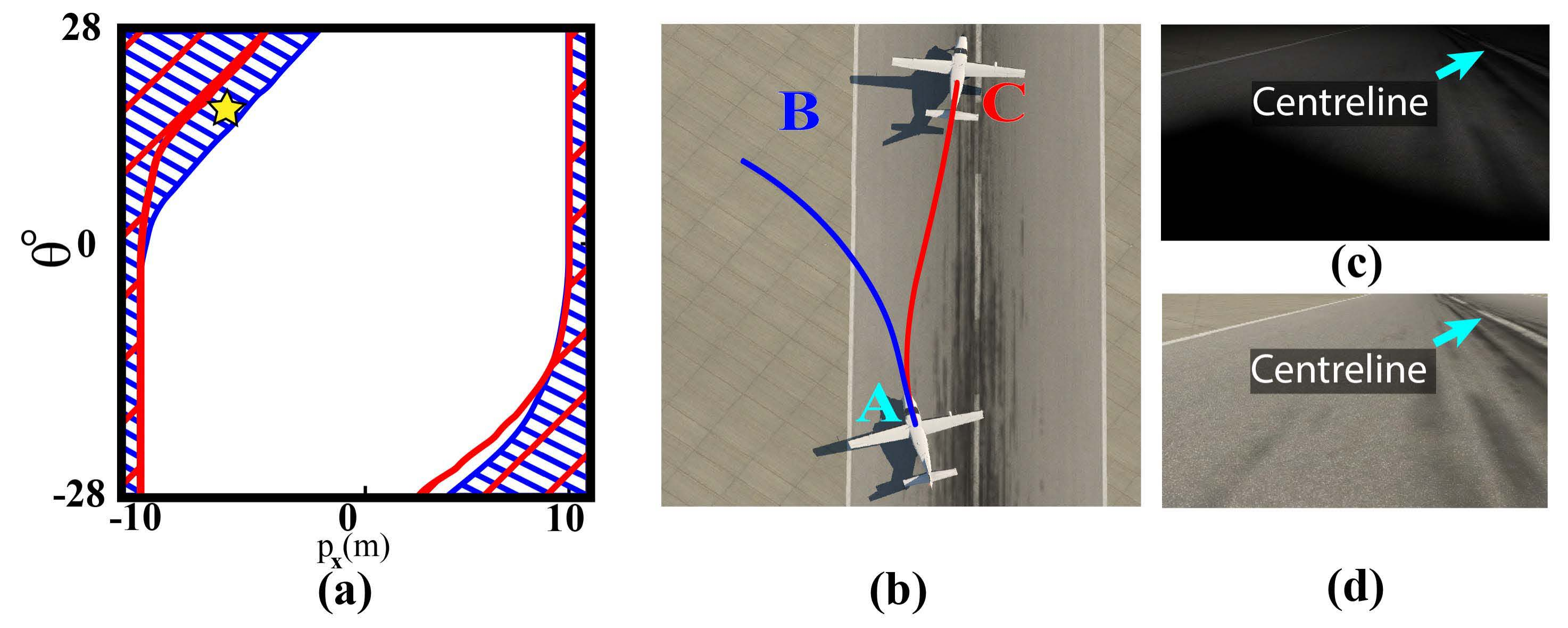}
     \vspace{-1.5em}
     \caption{
     \textbf{(a)} The morning (red shaded) and night (blue shaded) BRTs overlaid for $p_y$ = 190m. The state, shown with a yellow star, is only included in the night BRT. \textbf{(b)} Top view of the runway. In the morning, the CNN policy accomplishes the taxiing task by taking the red trajectory from ``A" (yellow star in (a)) to ``C." At night, the policy takes the aircraft outside the runway along the blue trajectory from ``A" to ``B". \textbf{(c)} The centreline in the image cannot be vividly seen by the CNN at location ``A" at night due to poor illumination, whereas it can be seen clearly in the morning \textbf{(d)}.}
     \label{fig:centrelane_fail_BRT}
     \vspace{-1.25em}
 \end{figure}

Comparing the BRTs for the morning and the night for other $p_y$, we observe that the night BRT is particularly big around the states where the aircraft starts near the left boundary of the runway (states around the top left corner of the Fig. \ref{fig:centrelane_fail_BRT}(a)). 

Based on the overlaid BRTs, we can predict that when the aircraft starts around the state marked by the yellow star, it should enter an unsafe state when guided by the CNN policy in the night-time but will successfully complete the taxiing task in the morning.
The observed images along the aircraft trajectory (Fig. \ref{fig:centrelane_fail_BRT}(c),(d)) expose that at night time the CNN is indeed unable to properly see the centreline due to illumination issues guiding the aircraft off the runway (blue trajectory from location A to B in Fig. \ref{fig:centrelane_fail_BRT}(b)). However, such errors are avoided in the morning (red trajectory from location A to C in Fig. \ref{fig:centrelane_fail_BRT}(b)) due to a better visibility. 

\noindent \textbf{{Comparison with forward-simulation methods.}} Our method took $\thicksim$6.5 hours to compute $\hat\policy$, the BRT, and isolate the failure cases for the entire system. 
The majority of this time was spent in computing $\hat\policy$, which involves rendering images at state grid points and querying the CNN. For comparison purposes, we also estimated the time that could be taken to find failure cases using a forward simulation of trajectory from each of the states in our state-space grid. 
The approximate computation time is around $\thicksim$67.5 days (it takes an average of 6s per trajectory simulation on a 24GB Nvidia RTX 3090 GPU. We required $\sim10^5$ input datapoints to sample the statespace appreciably.). 
The bottleneck in forward simulation primarily comes from a combination of \textit{repeated} image rendering, forward CNN calls, and post-processing of the CNN predictions for the trajectory simulation.
On the other hand, our reachability-based approach leverages dynamic programming to alleviate the need for repeated system queries.

Another advantage of our approach is its ability to expose hard-to-find, corner cases.
Corner cases, such as the failure due to the presence of a runway marking, occur from precise, sparsely-located starting positions in the aircraft's state space. 
In this case, it required us to complete an average of 145 simulations just to arrive at a single occurrence of such a failure mode when using a forward simulation-based random search.
The number of simulations is only expected to increase further with the number of robot states.
Finally, overlaying the BRT slices provides an intuitive way to compare the failures modes of a system across different environmental conditions.
These evaluations show the strength of HJ reachability-based methods in systematically discovering the closed-loop failures of high-dimensional, vision-based controllers over random or uniform grid searches.

\subsection{Autonomous Visual Navigation in Indoor Environments}
Our second case study analyzes a state-of-the-art visual navigation controller \cite{bansal2020combining} for a wheeled robot navigating an unknown indoor environment. 
The robot is modeled as a 5-dimensional nonlinear system with dynamics:
\vspace{-0.6em}
\begin{align*}
\label{eqn:dyn_lbwpn}
        \dot p_x = v cos(\theta),~\dot p_y = v sin(\theta),~\dot \theta = \omega,~\dot v = a,~\dot \omega = \alpha
        \vspace{-3em}
\end{align*}
Here, $p_x$ and $p_y$ denote the xy-position of the robot, $\theta$ is the robot heading, and $v \text{ and } \omega$ are its linear and angular speeds. It is assumed that the robot has access to perfect state estimates.
We define the control as $\ctrl \coloneqq (a,\ \alpha)$, where $a$ is the linear acceleration, and $\alpha$ is the angular acceleration. The navigation pipeline includes the CNN $C$ that accepts a $224\times224\times3$ RGB image $\image$ (from an onboard camera $\sensor$), the instantaneous linear and angular speeds of the robot ($v$ and $\omega$ respectively), and a goal position, $g$ to predict an intermediate waypoint position towards which the robot should move.
Finally, a model-based spline planner $P$ takes in the predicted waypoint to produce a smooth control profile for the robot. Hence, the closed-loop policy $\hat \policy$ is given by $\hat \policy :=  P\circ C \circ \sensor(\state,g,\env)$.
The environment $\env$ consists of various buildings in the Stanford Building Parser Dataset (SBPD) \cite{SBPD}. 

 \begin{figure}[h!]
      \centering
      \vspace{-0.7em}
     \includegraphics[width=0.85\columnwidth]{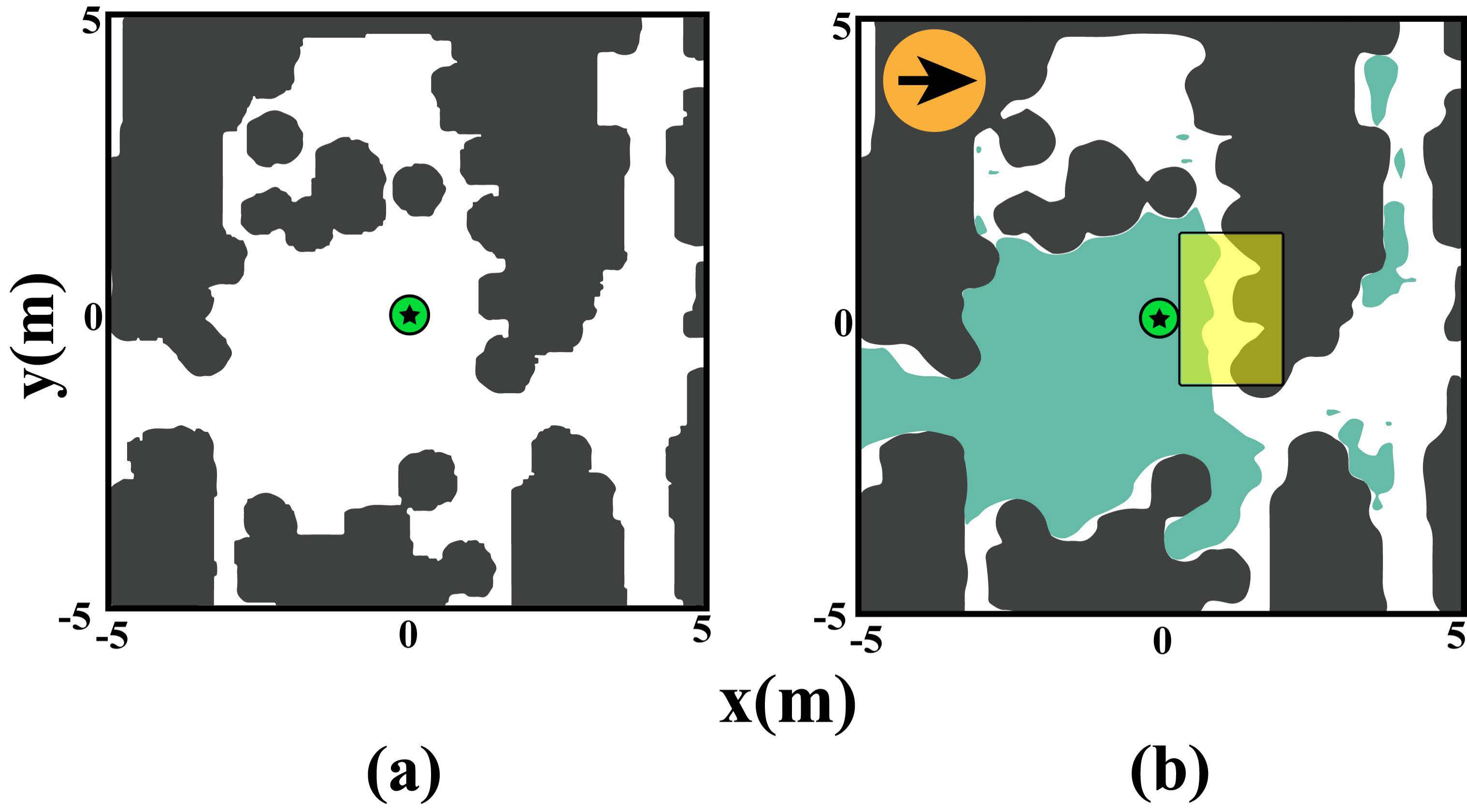}
     \vspace{-0.5em}
    \caption{\textbf{(a)} Ground truth occupancy map over a $5\times5$ area of an office space.
    The obstacles are shown in dark grey, the free space is shown in white, and the goal area is shown in green.
    \textbf{(b)} The corresponding BRAT slice is shown in teal for $\theta = 0\; rad$ (the initial heading is marked with the arrow inside the orange circle at the top left corner of the plot) with, $v = 0.3\; m/s$, and $\omega = 0\;rad/s$. 
    The yellow-shaded region highlights the states that are not included in the BRAT, indicating unsafe closed-loop behavior while starting from these states. 
    }
     \vspace{-0.75em}
     \label{fig:wpnav_all_brat}
 \end{figure}
Given a robot state, we can generate photo-realistic images ($\image = \sensor(\state,g,\env)$) of indoor scenes taken by a virtual monocular RGB camera $\sensor$, mounted on the robot. 
Representative top-view occupancy map from the environment is shown in Fig. \ref{fig:wpnav_all_brat}(a). Note that the obstacles are not known to the robot and it must navigate only using the first-person RGB images.

The CNN is a modified ResNet-50 architecture that is trained entirely in simulation using the SBPD dataset and shown to have a zero-shot sim-to-real generalization \cite{bansal2020combining}.
We wish to find and analyze the $\ibrt$ for the vision-based controller ($\hat \policy$) that leads to robot failures. 
In this case, failure is defined as the robot colliding with the obstacles, such as furniture or walls, before it reaches its goal. 
To find $\ibrt$, we compute the Backward Reach-Avoid Tube (BRAT) for the system -- the set of states from which the robot reaches its goal without colliding with any obstacles. 
The complement of the BRAT thus represents the unsafe states for the robot under $\hat \policy$.
BRAT can be computed by solving the HJB-VI in a similar fashion as BRT, with an additional constraint that the robot trajectory terminates once it reaches the obstacles.

The BRAT is computed for a 5-dimensional grid, of $51\times51\times21\times21\times21$ points, over a section of the robot's statespace. 
$\obs$ is given by the set of all obstacles in the environment, and the goal area is given by the positions that are within a distance of $0.3m$ from the position $g$.
For each state on the 5D grid, we render the observed RGB image using the SBPD simulator, query the waypoint using the CNN, and then obtain the control input using the planner $P$.
In Fig. \ref{fig:wpnav_all_brat}(b), we show the 2D projection of the 5D BRAT corresponding to the occupancy map in Fig. \ref{fig:wpnav_all_brat}(a).
If the robot starts from any state enclosed by the BRAT (teal area), it will reach the goal area (green circle) under the CNN-policy while avoiding the obstacles (shown in grey). 
Conversely, the area outside the BRAT represents the unsafe states for the robot.
Note that in this case, we do not compute the BRAT for the ideal system because the system is fully controllable, resulting in the ideal BRAT being simply the complement of the obstacle set.
\begin{figure}[h!]
      \centering
      \vspace{-0.75em}
     \includegraphics[width=0.95\columnwidth]{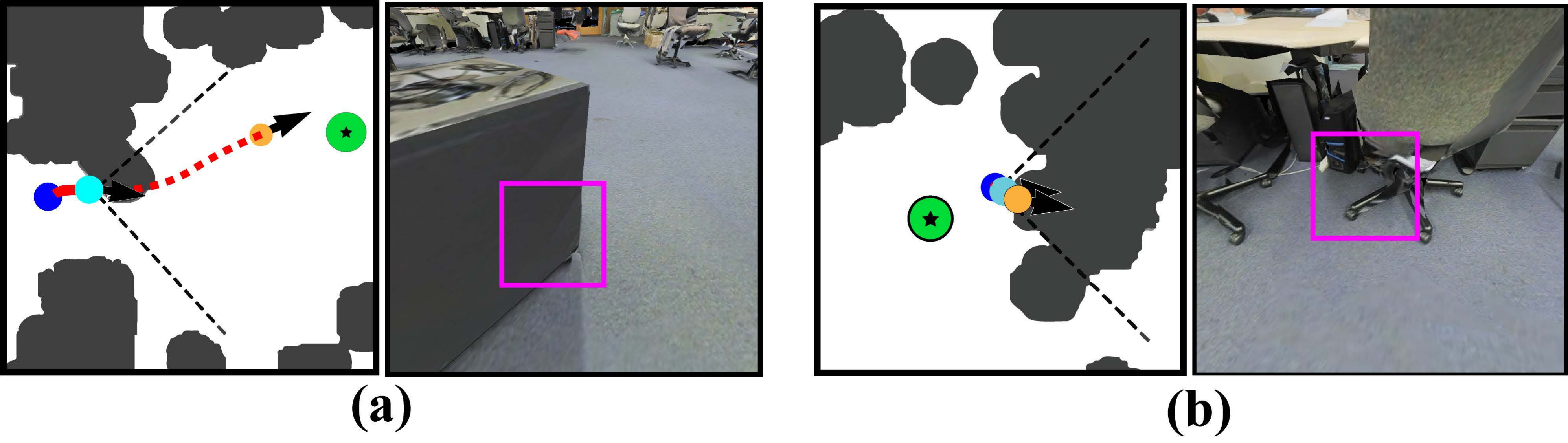}
    \caption{Simulation cases showing that the robot \textbf{(a)} collides with the corner of a table, or \textbf{(b)} runs head-on into the chairs. 
    The corresponding robot's trajectories are shown in (red). 
    The cyan marker shows the critical state of the robot. The field-of-view is shown with two dotted black lines from the critical state. The corresponding RGB image is the robot's observation at that state.
    The orange marker and the dotted red line respectively denote the predicted robot waypoint and trajectory at the critical state.
    The magenta square in the image highlights the collision region.
    }
     \vspace{-1.5em}
     \label{fig:precise_wp}
 \end{figure}

\noindent \textbf{Failure to predict precise waypoints near obstacles.} 
We notice that closer to the obstacles, the BRAT boundary is further from the obstacle boundary (e.g., yellow shaded region in Fig. \ref{fig:wpnav_all_brat}(b)), indicating that the vision-based controller leads to a collision when the robot is near the obstacles.
These cases are simulated in Figs. \ref{fig:precise_wp}(a), and (b).
Even though the closed-loop controller is trying to avoid the obstacles (the predicted robot trajectory is going around the obstacles), the trajectory is not turning enough and the robot collides with areas such as the corner of a table (Fig. \ref{fig:precise_wp}(a)), or the legs of a chair (Fig. \ref{fig:precise_wp}(b)).
Intuitively, the robot needs to predict very precise waypoints when it is closer to the obstacles, since even minor errors can lead to collisions.

\begin{wrapfigure}{r}{0.6\columnwidth}
          \centering
          \vspace{-1em}
         \includegraphics[width=0.6\columnwidth]{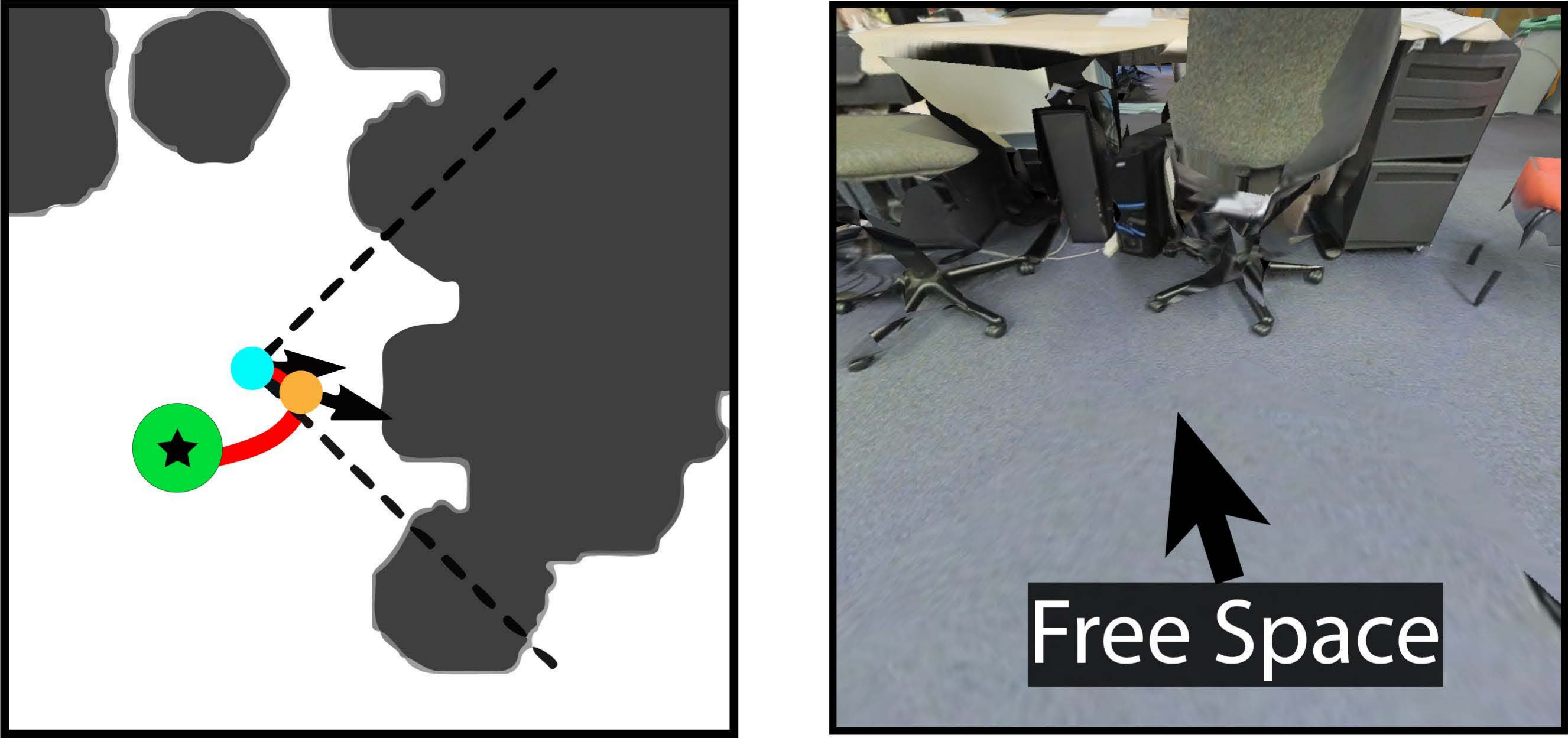}
         \caption{
         Allowing the robot visibility of the free space leads to a better waypoint prediction taking the robot to the goal.}
         \vspace{-1.0em}
         \label{fig:case_1_ablation}
\end{wrapfigure}%
However, it is challenging for the CNN to reason about the exact geometry of these complex obstacles solely from the first-person RGB images, which are particularly occluded near the obstacles, leading to inevitable errors in the waypoint prediction.

To confirm the CNN's reduction in performance near obstacles, we moved the robot's starting position away from the chairs for the simulation in fig. \ref{fig:precise_wp}(b). The CNN could then see a less occluded image (Fig. \ref{fig:case_1_ablation} right) and predicted a better waypoint to eventually reach the goal (Fig. \ref{fig:case_1_ablation} left).
     
\noindent \textbf{Failure to discern the traversability of narrow gaps.} 
Further analyzing the BRAT slices and the unsafe states indicate that the closed-loop controller tries to steer the robot through gaps that are too narrow for it to traverse through, given its geometry (a circular base with radius 0.2m). 
\begin{figure}[h!]
          \centering
         \includegraphics[width=0.85\columnwidth]{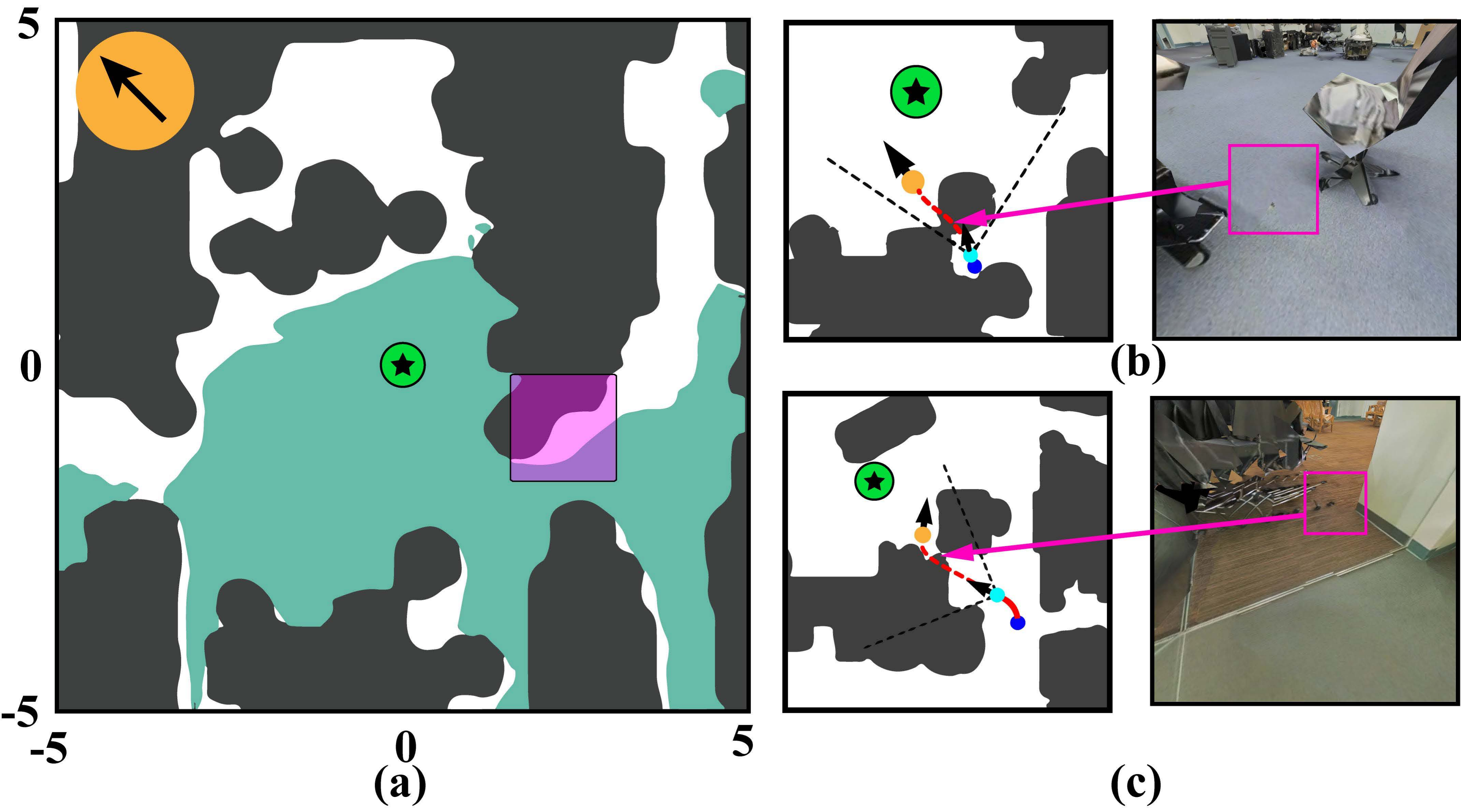}
         \caption{\textbf{(a)} The BRAT slice corresponding to $\theta = 3\pi/4$, $v = 0.3\; m/s$, and $\omega = 3\;rad/s$. The magenta square shows the states that are not covered by the BRAT. \textbf{(b)} Simulation from one such state shows that the CNN fails to discern the traversability of narrow passages. The CNN predicts a waypoint that leads the robot through the space between the chair and the table even though it does not possess enough clearance, eventually leading to a collision. The narrow passage is marked with a magenta square on the RGB image seen by the robot (right). \textbf{(c)} A similar case is observed between a stack of chairs and the wall.
         }
         \label{fig:case_2_narrow_path}
         \vspace{-1.5em}
     \end{figure}
Such states are shown in Fig. \ref{fig:case_2_narrow_path}(a) (inside the magenta-shaded region), and the robot trajectories from such states are simulated in Fig. \ref{fig:case_2_narrow_path}(b),(c).
Intuitively, from the RGB images, it seems that the gap is traversable, leading a RGB input-based CNN to predict a waypoint through the gap in order to reach the goal faster.
However, the gap is actually not traversable (as indicated by the absence of the gap in the occupancy map which is expanded by the size of the robot), ultimately leading to a collision with the obstacles.

Even when the gap is traversable, the CNN seems to struggle with steering the robot through narrow passages, as indicated by the unsafe states near the top of Fig. \ref{fig:case_2_narrow_path}(a).
These states ultimately need to go through a narrow passage in order to reach the goal which the closed-loop controller is unable to do reliably.
Such failures also indicate the shortcomings of the choice of sensor for the system.
For example, we hypothesize that adding another layer of depth estimation to the inputs could alleviate these failure modes, as it will provide the robot with better traversability information.
Targeted training of CNN near the narrow passages might also improve the robot performance in such scenarios.

\noindent \textbf{Misunderstanding certain obstacles as traversable.} Our BRAT slices indicate that the robot is able to traverse through hallways reasonably well; however, sometimes, it fails. 
\begin{figure}[h!]
      \centering
           \vspace{-1em}
     \includegraphics[width=0.9\columnwidth]{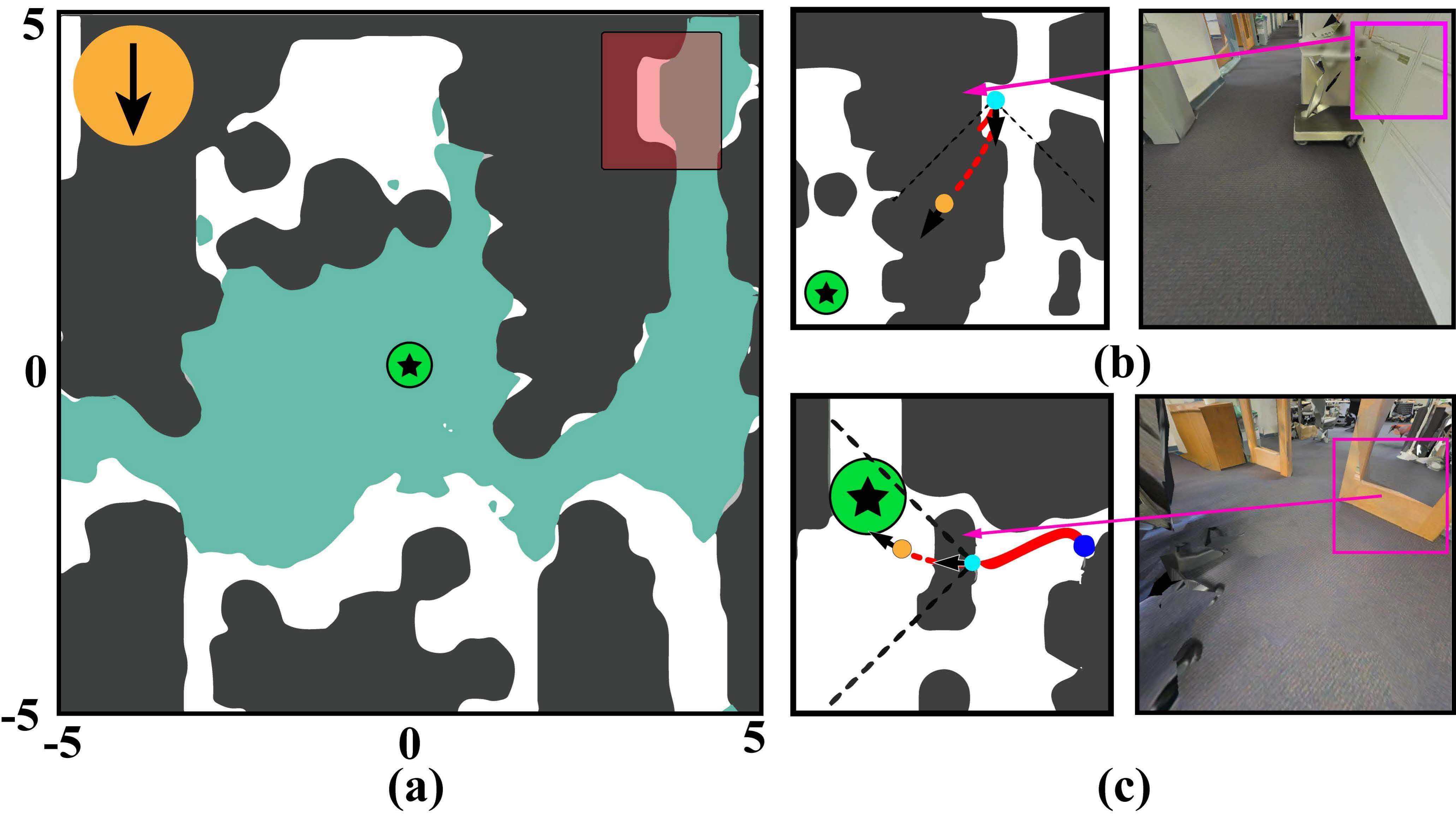}
     \caption{
     \textbf{(a)} Notice the highlighted area in the top-right location of the BRAT  for the robot heading of $-\pi / 2$ radians. Even though the robot faces down (wrt the top view), it cannot escape from the recessed region. \textbf{(b)} On simulating the robot  from one of the highlighted states, we saw that the CNN predicts a waypoint into the wall to its right and crashes the robot. We show the specific wall and its corresponding location on the top view with the magenta arrow. \textbf{(c)} Another situation  was observed where the robot crashed into a glass door due to the low height of the wooden pane around it. We show the glass door and its corresponding location on the top view with the magenta arrow.}
     \label{fig:case_3}
     \vspace{-1.em}
 \end{figure}
We highlight such states in the red-shaded region in Fig. \ref{fig:case_3}(a).
This is surprising since the robot simply needs to continue to move straight in such cases.
Simulating the robot trajectory (Fig. \ref{fig:case_3}(b)) from these states indicates that the CNN confidently predicts the waypoints inside the wall in order to reach the goal faster, ultimately leading to a collision.

\begin{wrapfigure}{r}{0.6\columnwidth}
          \centering
         \includegraphics[width=0.6\columnwidth]{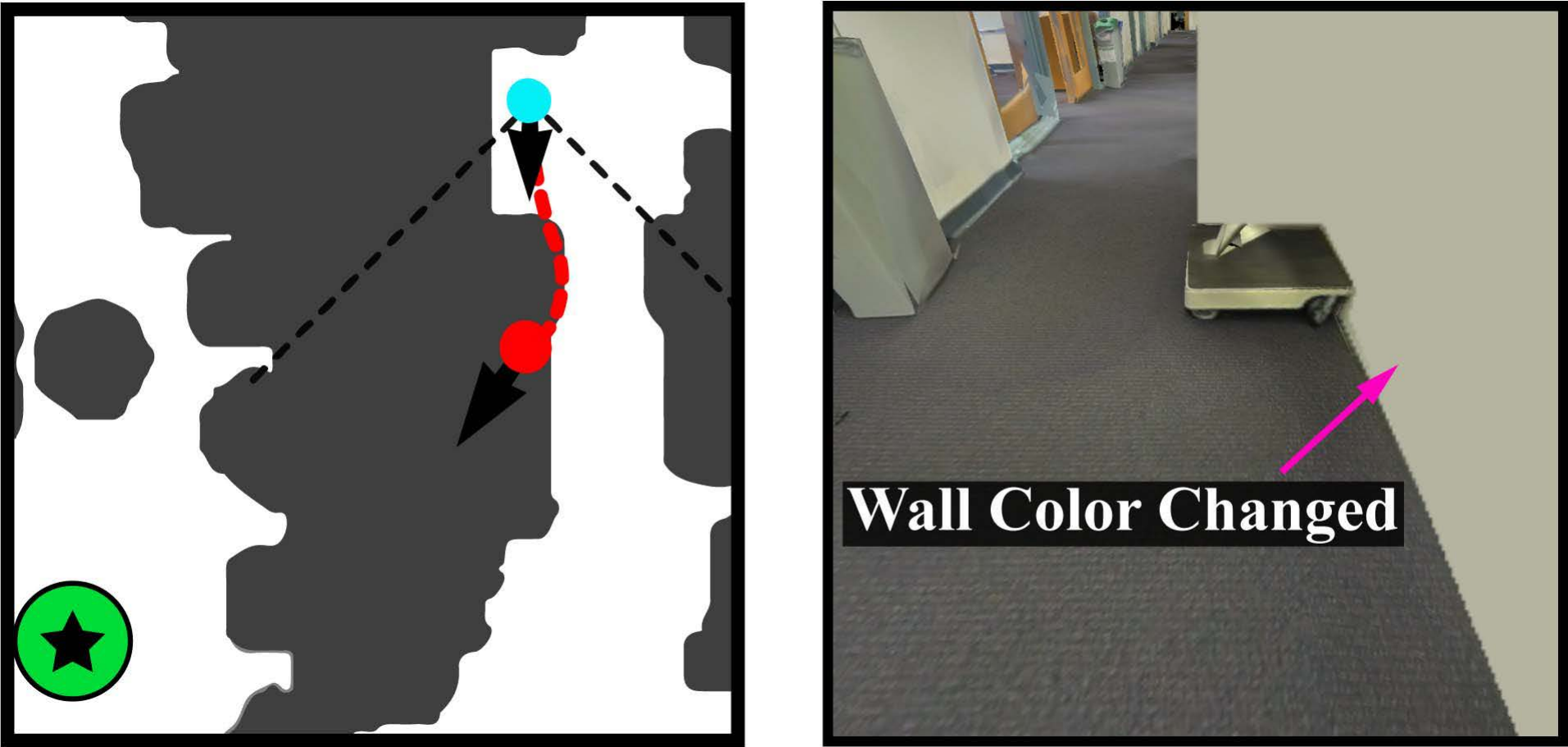}
         \caption{\textbf{(a)} On changing the color of the wall (right image), the CNN predicts a more feasible waypoint (left).}
         \vspace{-1em}
         \label{fig:case_3_ablation}
\end{wrapfigure}

To understand why the robot fails near this wall but not others, we change the color of the wall with the color picked from a different wall, as shown in the right image in Fig. \ref{fig:case_3_ablation}(a). This resulted in CNN correctly predicting the waypoint near the floor areas (left image in Fig. \ref{fig:case_3_ablation}(a)), indicating that the CNN somehow assumed the original wall to be traversable.
For our second example, we observed that if the CNN were fed with images containing obstacles with a small apparent height (such as glass gates), it would often ignore the obstacle as if it were a traversable area. 
A simulation at one of these failure states exposed a glass door, as shown in fig. \ref{fig:case_3}(c). The network conveniently ignores the glass door as it can see the ground area through it. 

\noindent \textbf{{Comparison with forward-simulation methods.}} 
Our method took $\thicksim$6.5 days to compute the BRAT, out of which 6 days were spent on image rendering at grid points and the CNN query.
Even though the computation time is non-trivial, it is still significantly lower than that for a forward simulation-based procedure, which is estimated to be $\thicksim$11K days (extrapolated based on 10 trajectories). We recorded an average simulation time  of $\sim40$ seconds per trajectory over a statespace discretized into $\sim2.4\times10^7$ datapoints.
Most of this computation time is again due to the repeated queries of the simulator, forward passes through the CNN, and post-processing the CNN predictions while simulating a trajectory, which can be extremely slow.

Moreover, the BRAT slices provide an intuitive clustering of the failure states, which can be helpful for targeted data collection to improve the CNN. 

\section{Discussion and Future Work}
\label{sec:conclusion}
We introduce a framework for automatically discovering the closed-loop failures of vision-based controllers. 
Our work combines simulation-based approaches with HJ reachability analysis to systematically and tractably find these failures. 
We demonstrate the efficacy of our approach on two distinct applications -- a 5-dimensional wheeled robot and a 3-dimensional aircraft using RGB-image-based neural network controllers. 

Our work suggests a number of interesting future directions. 
First, the grid-based approach to computing reachable sets suffers from the curse of dimensionality. 
In the future, it would be interesting to consider alternative sampling-based reachability methods, such as DeepReach \cite{bansal2021deepreach}, that are shown to scale well to high-dimensional systems.
We will also like to explore using the obtained failures in improving the performance of the vision-based controller, e.g., through incremental training on the obtained scenarios.
Finally, even though the failure discovery is automatic, the analysis of the failures is primarily performed manually in the current work.
Automating the failure analysis, e.g., through generative and clustering methods, will be a promising future direction.

\bibliographystyle{IEEEtran}
\bibliography{references}
\end{document}